\documentclass[11pt,a4paper]{article}
\usepackage[hyperref]{acl2019}
\usepackage{times}
\usepackage{latexsym}

\usepackage{subcaption}
\usepackage{graphicx}

\usepackage{url}
\usepackage{lipsum}

\usepackage[normalem]{ulem} 
\usepackage{soul,color}
\usepackage{todonotes}
\usepackage{booktabs}

\aclfinalcopy 

\title{Improving Neural Language Models by\\ Segmenting, Attending, and Predicting the Future}

\author{Hongyin Luo$^1$ $    $ Lan Jiang$^2$ $    $ Yonatan Belinkov$^1$ $    $ James Glass$^1$ \\
  $^{1}$MIT Computer Science and Artificial Intelligence Laboratory, Cambridge, MA 02139, USA \\
  {\tt \{hyluo, belinkov, glass\}@mit.edu}\\
  $^{2}$School of Information Sciences, University of Illinois at Urbana--Champaign\\ Champaign, IL
61820, USA \\ 
  {\tt lanj3@illinois.edu}
  }

\date{}

\begin{document}
\maketitle
\begin{abstract}
Common language models typically predict the next word given the context. In this work, we propose a method that improves language modeling by learning to align the given context and the following phrase. The model does not require any linguistic annotation of phrase segmentation. Instead, we define syntactic heights and phrase segmentation rules, enabling the model to automatically induce phrases, recognize their task-specific heads, and generate phrase embeddings in an unsupervised learning manner. Our method can easily be applied to language models with different network architectures since an independent module is used for phrase induction and context-phrase alignment, and no change is required in the underlying language modeling network. Experiments have shown that our model outperformed several strong baseline models on different data sets. We achieved a new state-of-the-art performance of 17.4 perplexity on the Wikitext-103 dataset. Additionally, visualizing the outputs of the phrase induction module showed that our model is able to learn approximate phrase-level structural knowledge without any annotation.
\end{abstract}

\section{Introduction}

Neural language models
are typically trained
by predicting the next word given a past context \cite{bengio2003neural}. However, natural sentences are not constructed as simple linear word sequences, as they usually contain complex syntactic information. For example, a subsequence of words can constitute a phrase, and two non-neighboring words can depend on each other. These properties make natural sentences more complex than simple linear sequences.

Most recent work on neural language modeling learns a model by encoding contexts and matching the context embeddings to the embedding of the next word \cite{bengio2003neural,merity2017regularizing,melis2017state}. In this line of work, a given context is encoded with a neural network, for example a long short-term memory \citep[LSTM;][]{hochreiter1997long} network, and is represented with a distributed vector. The log-likelihood of predicting a word is computed by calculating the inner product between the word embedding and the context embedding. Although most models do not explicitly consider syntax, they still achieve state-of-the-art performance on different corpora. Efforts have also been made to utilize structural information to learn better language models. For instance, parsing-reading-predict networks \citep[PRPN;][]{shen2017neural} explicitly learn a constituent parsing structure of a sentence and predict the next word considering the internal structure of the given context with an attention mechanism. Experiments have shown that the model is able to capture some syntactic information.

Similar to word representation learning models that learns to match word-to-word relation matrices \cite{mikolov2013distributed,pennington2014glove}, standard language models are trained to factorize context-to-word relation matrices \cite{yang2017breaking}. In such work, the context comprises all previous words observed by a model for predicting the next word. However, we believe that context-to-word relation matrices are not sufficient for describing how natural sentences are constructed.
We argue that natural sentences are generated at a higher level before being decoded to words. Hence a language model should be able to predict the following sequence of words given a context. In this work, we propose a model that factorizes a context-to-phrase mutual information matrix to learn better language models. The context-to-phrase mutual information matrix describes the relation among contexts and the probabilities of phrases following given contexts. We make the following contributions in this paper:

\begin{itemize}
\item We propose a phrase prediction model that improves the performance of state-of-the-art word-level language models. 
\item Our model learns to predict approximate phrases and headwords without any annotation.
\end{itemize}

\section{Related Work}

Neural networks have been widely applied in natural language modeling and generation \cite{bengio2003neural,bahdanau2014neural} for both encoding and decoding. Among different neural architectures, the most popular models are recurrent neural networks  \citep[RNNs;][]{mikolov2010recurrent}, long short-term memory networks  \citep[LSTMs;][]{hochreiter1997long}, and convolutional neural networks  \cite[CNNs;][]{bai2018empirical,dauphin2017language}.

Many modifications of network structures have been made based on these architectures. LSTMs with self-attention can improve the performance of language modeling \cite{tran2016recurrent,cheng2016long}. As an extension of simple self-attention, transformers \cite{vaswani2017attention} apply multi-head self-attention and have achieved competitive performance compared with recurrent neural language models. A current state-of-the-art model, Transformer-XL \cite{dai2018transformer}, applied both a recurrent architecture and a multi-head attention mechanism. To improve the quality of input word embeddings, character-level information is also considered \cite{kim2016character}. It has also been shown that context encoders can learn syntactic information \cite{shen2017neural}.

However, instead of introducing architectural changes, for example a self-attention mechanism or character-level information, previous studies have shown that careful hyper-parameter tuning and regularization techniques on standard LSTM language models can obtain significant improvements \cite{melis2017state,merity2017regularizing}. Similarly, applying more careful dropout strategies can also improve the language models \cite{gal2016theoretically,melis2018pushing}. LSTM language models can be improved with these approaches because LSTMs suffer from serious over-fitting problems.

Recently, researchers have also attempted to improve language models at the decoding phase. \citet{inan2016tying} showed that reusing the input word embeddings in the decoder can reduce the perplexity of language models. \citet{yang2017breaking} showed the low-rank issue in factorizing the context-to-word mutual information matrix and proposed a multi-head softmax decoder to solve the problem. Instead of predicting the next word by using only similarities between contexts and words, the neural cache model \cite{grave2016improving} can significantly improve language modeling by considering the global word distributions conditioned on the same contexts in other parts of the corpus.

To learn the grammar and syntax in natural languages, \citet{dyer2016recurrent} proposed the recurrent neural network grammar (RNNG) that models language incorporating a transition parsing model. Syntax annotations are required in this model. To utilize the constituent structures in language modeling without syntax annotation, parse-read-predict networks  \citep[PRPNs;][]{shen2017neural} calculate syntactic distances among words and computes self-attentions. Syntactic distances have been proved effective in constituent parsing tasks \cite{shen2018straight}. In this work, we learn phrase segmentation with a model based on this method and our model does not require syntax annotation.




\section{Syntactic Height and Phrase Induction}
\label{sec:psc}
In this work, we propose a language model that not only predicts the next word of a given context, but also attempts to match the embedding of the next phrase. The first step of this approach is conducting phrase induction based on syntactic heights. In this section, we explain the definition of syntactic height in our approach and describe the basics ideas about whether a word can be included in an induced phrase.

\begin{figure}[t]
\centering
\includegraphics[height=1.8in]{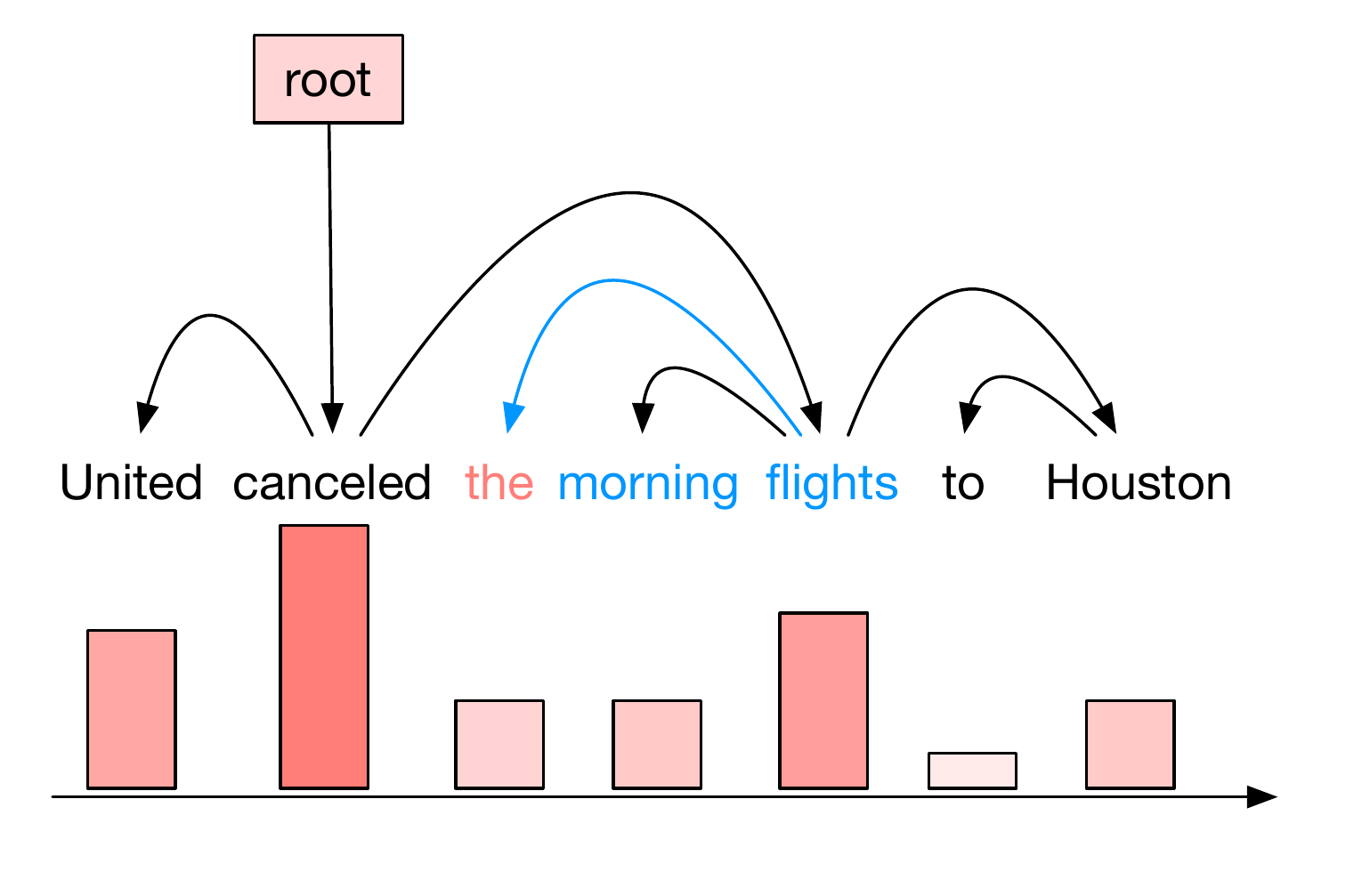}
\caption{Groundtruth dependency tree and syntactic heights of each word.}
\label{fig:dt}
\end{figure}

Intuitively, the syntactic height of a word aims to capture its distance to the root node in a dependency tree. In Figure \ref{fig:dt}, the syntactic heights are represented by the red bars. A word has high syntactic height if it has low distance to the root node.

A similar idea, named syntactic distance, is proposed by \citet{shen2017neural} for constructing constituent parsing trees. We apply the method for calculating syntactic distance to calculate syntactic height. Given a sequence of embeddings of input words $ [x_1, x_2, \cdots, x_n] $, we calculate their syntactic heights with a temporal convolutional network (TCN) \cite{bai2018empirical}. 
%
%
\begin{equation}
d_i = W_d \cdot [x_{i - n}, x_{i - n + 1}, \cdots, x_i]^T + b_d
\end{equation}
\begin{equation}
h_i = W_h \cdot ReLU(d_i) + b_h
\end{equation}
where $h_i$ stands for the syntactic height of word $x_i$. The syntactic height $h_i$ for each word is a scalar, and $W_h$ is a $1 \times D$ matrix, where $D$ is the dimensionality of $d_i$. These heights are learned and not imposed by external syntactic supervision. 
In \citet{shen2017neural}, the syntactic heights are used to generate context embeddings. In our work, we use the syntactic heights to predict induced phrases and calculate their embeddings.

We define the phrase induced by a word based on the syntactic heights. Consider two words $x_i$ and $x_k$. $x_k$ belongs to the phrase induced by $x_i$ if and only if for any $j \in (i, k)$, $h_j < max(h_i, h_k)$. For example, in Figure \ref{fig:dt}, the phrase induced by the red marked word \textbf{the} is ``the morning flights'', since the syntactic height of the word \textbf{morning}, $h_{morning} < h_{flights}$. However, the word ``to'' does not belong to the phrase because $h_{flights}$ is higher than both $h_{the}$ and $h_{to}$. The induced phrase and the inducing dependency connection are labeled in blue in the figure.

Note that this definition of an induced phrase does not necessarily correspond to a phrase in the syntactic constituency sense. For instance, the words ``to Houston'' would be included in the phrase ``the morning flights to Houston'' in a traditional syntactic tree. Given the definition of induced phrases, we propose phrase segmenting conditions (PSCs) to find the last word of an induced phrase. Considering the induced phrase of the $i$-th word, $s_i = [x_i, x_{i + 1}, \cdots, x_j]$. If $x_j$ is not the last word of a given sentence, there are two conditions that $x_j$ should satisfy:
\begin{itemize}
\item[1.] (PSC-1) The syntactic height of $x_j$ must be higher than the height of $x_i$, that is
\begin{equation}
h_j - h_i > 0
\end{equation}
\item[2.] (PSC-2) The syntactic height of $x_{j+1}$ should be lower that $x_j$.
\begin{equation}
h_j - h_{j+1} > 0
\end{equation}
\end{itemize}

Given the PSCs, we can decide the induced phrases for the sentence shown in Figure \ref{fig:dt}. The last word of the phrase induced by ``United'' is ``canceled'', and the last word of the phrase induced by ``flights'' is ``Houston''. For the word assigned the highest syntactic height, its induced phrase is all remaining words in the sentence.

\section{Model}
In this work, we formulate multi-layer neural language models as a two-part framework. For example, in a two-layer LSTM language model \cite{merity2017regularizing}, we use the first layer as phrase generator and the last layer as a word generator:
%
%
\begin{equation}
\label{eq:c}
[c_1, c_2, \cdots, c_T] = RNN^1 ([x_1, x_2, \cdots, x_T])
\end{equation}
\begin{equation}
[y_1, y_2, \cdots, y_T] = RNN^2 ([c_1, c_2, \cdots, c_T])
\end{equation}

For a $L$-layer network, we can regard the first $L_1$ layers as the phrase generator and the next $L_2 = L - L_1$ layers as the word generator. Note that we use $y_i$ to represent the hidden state output by the second layer instead of $h_i$, since $h_i$ in our work is defined as the syntactic height of $x_i$. In the traditional setting, the first layer does not explicitly learn the semantics of the following phrase because there is no extra objective function for phrase learning.

\begin{figure*}[t]
\centering
\includegraphics[height=2.1in]{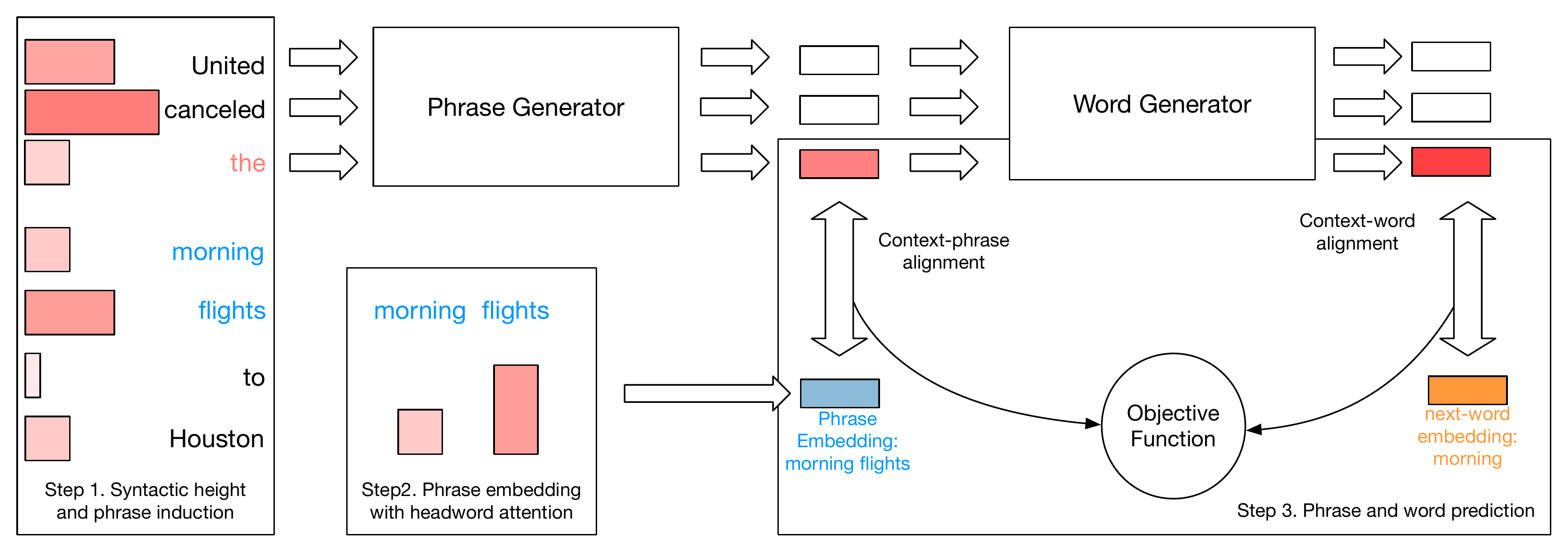}
\caption{The 3-step diagram of our approach. The current target word is ``the'', the induced phrase is ``morning flights'', and the next word is ``morning''. The context-phrase and context-word alignments are jointly trained.}
\label{fig:diagram}
\end{figure*}

In this work, we force the first layer to output context embeddings $c_i$ for phrase prediction with three steps. Firstly, we predict the induced phrase for each word according to the PSCs proposed in Section \ref{sec:psc}. Secondly, we calculate the embedding of each phrase with a head-finding attention. Lastly, we align the context embedding and phrase embedding with negative sampling. The word generation is trained in the same way as standard language models. The diagram of the model is shown in Figure \ref{fig:diagram}.
The three steps are described next. 

\subsection{Phrase Segmentation}

We calculate the syntactic height and predict the induced phrase for each word:
\begin{equation}
h_i = TCN([x_{i - n}, x_{i - n + 1}, \cdots, x_i])
\end{equation}
where $TCN(\cdot)$ stands for the TCN model described in Equations (1) and (2), and $n$ is the width of the convolution window.

Based on the proposed phrase segmenting conditions (PSCs) described in the previous section, we predict the probability of a word being the first word outside a induced phrase. Firstly, we decide if each word, $x_{j-1}, j \in (i+1, n]$, satisfies the two phrase segmenting conditions, PSC-1 and PSC-2. The probability that $x_j$ satisfies PSC-1 is 
%
%
\begin{equation}
\label{eq:psc-1}
p_{psc}^1 (x_j) = \frac{1}{2} \cdot (f^{HT} (h_j - h_i) + 1)
\end{equation}
Similarly, the probability that $x_j$ satisfies PSC-2 is
\begin{equation}
\label{eq:psc-2}
p_{psc}^2 (x_j) = \frac{1}{2} \cdot (f^{HT} (h_j - h_{j+1}) + 1)
\end{equation}
where $f_{HT}$ stands for the HardTanh function with a temperature $a$:
$$ f^{HT} (x)=\left\{
\begin{array}{rcl}
-1       &      & {x      \leq      -\frac{1}{a}}\\
a \cdot x     &      & {-\frac{1}{a} < x \leq \frac{1}{a}}\\
1     &      & {x > \frac{1}{a}}
\end{array} \right. $$
This approach is inspired by the context attention method proposed in the PRPN model \cite{shen2017neural}.

Then we can infer the probability of whether a word belongs to the induced phrase of $x_i$ with 
\begin{equation}
\label{eq:p-ind}
p^{ind} (x_j) = \prod_{k=1}^j \hat{p}(x_k)
\end{equation}
where $p^{ind} (x_i)$ stands for the probability that $x_i$ belongs to the induced phrase, and 
$$ \scriptstyle \hat{p} (x_k)=\left\{
\begin{array}{rcl}
& 1             & {k      \leq      i+1}\\
& 1 - p_{psc}^1 (x_{k-1}) \cdot p_{psc}^2 (x_{k-1})           & {k > i+1}
\end{array} \right. $$
Note that the factorization in Equation~\ref{eq:p-ind} assumes that words are independently likely to be included in the induced phrase of $x_i$. 


\subsection{Phrase Embedding with Attention}

Given induced phrases, we can calculate their embeddings based on syntactic heights. To calculate the embedding of phrase $s = [x_1, x_2, \cdots, x_n]$, we calculate an attention distribution over the phrase:
\begin{equation}
\label{eq:att}
\alpha_i = \frac{h_i \cdot p^{ind} (x_i) + c}{\sum_j h_j \cdot p^{ind} (x_j) + c}
\end{equation}
where $h_i$ stands for the syntactic height for word $x_i$ and $c$ is a constant real number for smoothing the attention distribution. Then we generate the phrase embedding with a linear transformation:
\begin{equation}
s = W \cdot \sum_i \alpha_i \cdot e_i
\end{equation}
where $e_i$ is the word embedding of $x_i$. In training, we apply a dropout layer on $s$.

\subsection{Phrase and Word Prediction}

A traditional language model learns the probability of a sequence of words:
\begin{equation}
p(x_1, x_2, \cdots, x_n) = p(x_1) \cdot \prod_i p(x_{i+1} | x_1^{i})
\end{equation}
where $x_1^i$ stands for $x_1, x_2, \cdots, x_i$, which is the context used for predicting the next word, $x_{i+1}$. In most related studies, the probability $p(x_{i+1}|x_1^i)$ is calculated with the output of the top layer of a neural network $y_i$ and the word representations $e_{i+1}$ learned by the decoding layer:
\begin{equation}
p(x_{i+1}) = Softmax(e_{i+1}^T \cdot y_i)
\end{equation}

The state-of-the-art neural language models contain multiple layers. The outputs of different hidden layers capture different level of semantics of the context. In this work, we force one of the hidden layers to align its output with the embeddings of induced phrases $s_i$. We apply an embedding model similar to \citet{mikolov2013distributed} to train the hidden output and phrase embedding alignment. We define the context-phrase alignment model as follows.
%
%
%
%
%
%
%

We first define the probability that a phrase $ph_i$ can be induced by context $[x_1, \dots, x_i]$.
\begin{equation}
p(ph_i | x_1^i) = \sigma(c_i^T \cdot s_i)
\end{equation}
where $\sigma(x) = \frac{1}{1 + e^{-x}}$, and $c_i$ stands for the context embedding of $x_1, x_2, \cdots, x_i$ output by a hidden layer, defined in Equation \ref{eq:c}. $s_i$ is the generated embedding of an induced phrase. The probability that a phrase $ph_i$ cannot be induced by context $[x_1, \dots, x_i]$ is $1 - p(ph_i | x_1^i)$. This approach follows the method for learning word embeddings proposed in \citet{mikolov2013distributed}.

We use an extra objective function and the negative sampling strategy to align context representations and the embeddings of induced phrases. Given the context embedding $c_i$, the induced phrase embedding $s_i$, and random sampled negative phrase embeddings $s_i^{neg}$, we train the neural network to maximize the likelihood of true induced phrases and minimize the likelihood of negative samples. we define the following objective function for context $i$:
\begin{equation}
l_i^{CPA} = 1 - \sigma(c_i^T \cdot s_i) + \frac{1}{n} \sum_{j=1}^n \sigma (c_i^T \cdot s_j^{neg})
\end{equation}
where $n$ stands for the number of negative samples. With this loss function, the model learns to maximize the similarity between the context and true induced phrase embeddings, and minimize the similarity between the context and negative samples randomly selected from the induced phrases of other words. In practice, this loss function is used as a regularization term with a coefficient $\gamma$:
\begin{equation}
l = l^{LM} + \gamma \cdot l^{CPA}
\end{equation}

It worth noting that our approach is model-agnostic and and can be applied to various architectures. The TCN network for calculating the syntactic heights and phrase inducing is an independent module. In context-phrase alignment training with negative sampling, the objective function provides phrase-aware gradients and does not change the word-by-word generation process of the language model. 

\section{Experiments}

We evaluate our model with word-level language modeling tasks on Penn Treebank  \cite[PTB;][]{mikolov2010recurrent}, Wikitext-2  \cite[WT2;][]{bradbury2016quasi}, and Wikitext-103  \cite[WT103;][]{merity2016pointer} corpora.

The PTB dataset has a vocabulary size of 10,000 unique words. The entire corpus includes roughly 40,000 sentences in the training set, and more than 3,000 sentences in both valid and test set.

The WT2 data is about two times larger the the PTB dataset. The dataset consists of Wikipedia articles. The corpus includes 30,000 unique words in its vocabulary and is not cleaned as heavily as the PTB corpus.

The WT103 corpus contains a larger vocabulary and more articles than WT2. It consists of 28k articles and more than 100M words in the training set. WT2 and WT103 corpora can evaluate the ability of capturing long-term dependencies \cite{dai2018transformer}.


In each corpus, we apply our approach to publicly-available, state-of-the-art models. This demonstrates that our approach can improve different existing architectures. Our trained models will be published for downloading. The implementation of our models is publicly available.\footnote{\url{https://github.com/luohongyin/PILM}}

\begin{table*}[ht]
\centering
\begin{tabular}{p{0.5\linewidth}p{0.1\linewidth}p{0.1\linewidth}p{0.1\linewidth}}
\toprule
\textbf{Model} & \textbf{\#Params} & \textbf{Dev PPL} & \textbf{Test PPL}\\
\midrule
\citet{inan2016tying} -- Tied Variational LSTM & 24M & 75.7 & 73.2\\
\citet{zilly2017recurrent} -- Recurrent Highway Networks & 23M & 67.9 & 65.7\\
\citet{shen2017neural} -- PRPN & - & - & 62.0\\
\citet{pham2018efficient} -- Efficient NAS & 24M & 60.8 & 58.6\\
\citet{melis2017state} -- 4-layer skip LSTM (tied) & 24M & 60.9 & 58.3\\
\citet{shen2018ordered} -- ON-LSTM & 25M & 58.3 & 56.2\\
\citet{liu2018darts} -- Differentiable NAS & 23M & 58.3 & 56.1\\
\midrule
\citet{merity2017regularizing} -- AWD-LSTM & 24M & 60.7 & 58.8\\
\citet{merity2017regularizing} -- AWD-LSTM + finetuning & 24M & 60.0 & 57.3\\
\midrule
Ours -- AWD-LSTM + Phrase Induction - NS & 24M & 61.0 & 58.6 \\
Ours -- AWD-LSTM + Phrase Induction - Attention & 24M & 60.2 & 58.0 \\
Ours -- AWD-LSTM + Phrase Induction & 24M & 59.6 & 57.5\\
Ours -- AWD-LSTM + Phrase Induction + finetuning & 24M & 57.8 & 55.7 \\
\midrule
\citet{dai2018transformer} -- Transformer-XL & 24M & 56.7 & 54.5\\
\citet{yang2017breaking} -- AWD-LSTM-MoS + finetuning & 22M & \textbf{56.5} & \textbf{54.4}\\
\bottomrule
\end{tabular}
\caption{Experimental results on Penn Treebank dataset. Compared with the AWD-LSTM baseline models, our method reduced the perplexity on test set by 1.6.}
\label{tab:ptb}
\end{table*}

\begin{table*}[ht]
\centering
\begin{tabular}{p{0.5\linewidth}p{0.1\linewidth}p{0.1\linewidth}p{0.1\linewidth}}
\toprule
\textbf{Model} & \textbf{\#Params} & \textbf{Dev PPL} & \textbf{Test PPL}\\
\midrule
\citet{inan2016tying} -- Variational LSTM (tied) & 28M & 92.3 & 87.7\\
\citet{inan2016tying} -- VLSTM + augmented loss & 28M & 91.5 & 87.0\\
\citet{grave2016improving} -- LSTM & - & - & 99.3\\
\citet{grave2016improving} -- LSTM + Neural cache & - & - & 68.9\\
\citet{melis2017state} -- 1-Layer LSTM & 24M & 69.3 & 69.9\\
\citet{melis2017state} -- 2-Layer Skip Conn. LSTM & 24M & 69.1 & 65.9\\
\midrule
\citet{merity2017regularizing} -- AWD-LSTM + finetuning & 33M & 68.6 & 65.8\\
\midrule
Ours -- AWD-LSTM + Phrase Induction & 33M & 68.4 & 65.2\\
Ours -- AWD-LSTM + Phrase Induction + finetuning & 33M & \textbf{66.9} & \textbf{64.1}\\
\bottomrule
\end{tabular}
\caption{Experimental results on Wikitext-2 dataset.}
\label{tab:wt2}
\end{table*}

\begin{table*}[ht]
\centering
\begin{tabular}{p{0.5\linewidth}p{0.1\linewidth}p{0.1\linewidth}}
\toprule
\textbf{Model} & \textbf{\#Params} & \textbf{Test PPL}\\
\midrule
\citet{grave2016improving} -- LSTM & - & 48.7\\
\citet{bai2018empirical} -- TCN & - & 45.2\\
\citet{dauphin2017language} -- GCNN-8 & - & 44.9\\
\citet{grave2016improving} -- LSTM + Neural cache & - & 40.8\\
\citet{dauphin2017language} -- GCNN-14 & - & 37.2\\
\citet{merity2018analysis} -- 4-layer QRNN & 151M & 33.0\\
\citet{rae2018fast} -- LSTM + Hebbian + Cache & - & 29.9\\
\citet{dai2018transformer} -- Transformer-XL Standard & 151M & 24.0\\
\citet{baevski2018adaptive} -- Adaptive input & 247M & 20.5\\
\midrule
\citet{dai2018transformer} -- Transformer-XL Large & 257M & 18.3\\
Ours -- Transformer-XL Large + Phrase Induction & 257M & \textbf{17.4}\\
\bottomrule
\end{tabular}
\caption{Experimental results on Wikitext-103 dataset.}
\label{tab:wt103}
\end{table*}


\subsection{Penn Treebank}
We train a 3-layer AWD-LSTM language model \cite{merity2017regularizing} on PTB data set. We use 1,150 as the number of hidden neurons and 400 as the size of word embeddings. We also apply the word embedding tying strategy \cite{inan2016tying}. We apply variational dropout for hidden states \cite{gal2016theoretically} and the dropout rate is 0.25. We also apply weight dropout \cite{merity2017regularizing} and set weight dropout rate as 0.5. We apply stochastic gradient descent (SGD) and averaged SGD \citep[ASGD;][]{polyak1992acceleration} for training. The learning rate is 30 and we clip the gradients with a norm of 0.25. For the phrase induction model, we randomly sample 1 negative sample for each context, and the context-phrase alignment loss is given a coefficient of 0.5. The output of the second layer of the neural network is used for learning context-phrase alignment, and the final layer is used for word generation.

We compare the word-level perplexity of our model with other state-of-the-art models and our baseline is AWD-LSTM \cite{merity2017regularizing}.
The experimental results are shown in Table \ref{tab:ptb}. Although not as good as the Transformer-XL model \cite{dai2018transformer} and the mixture of softmax model \cite{yang2017breaking}, our model significantly improved the AWD-LSTM, reducing 2.2 points of perplexity on the validation set and 1.6 points of perplexity on the test set. Note that the ``finetuning'' process stands for further training the language models with ASGD algorithm \cite{merity2017regularizing}.

We also did an ablation study without either headword attention or negative sampling (NS). The  results are listed in Table \ref{tab:ptb}. By simply averaging word vectors in the induced phrase Without the attention mechanism, the model performs worse than the full model by 0.5 perplexity, but is still better than our baseline, the AWD-LSTM model. In the experiment without negative sampling, we only use the embedding of true induced phrases to align with the context embedding. It is also indicated that the negative sampling strategy can improve the performance by 1.1 perplexity. Hence we just test the full model in the following experiments.

\subsection{Wikitext-2}

We also trained a 3-layer AWD-LSTM language model on the WT2 dataset. The network has the same input size, output size, and hidden size as the model we applied on PTB dataset, following the experiments done by \citet{merity2017regularizing}. Some hyper-parameters are different from the PTB language model. We use a batch size of 60. The embedding dropout rate is 0.65 and the dropout rate of hidden outputs is set to 0.2. Other hyper-parameters are the same as we set in training on the PTB dataset.

The experimental results are shown in Table \ref{tab:wt2}. Our model improves the AWD-LSTM model by reducing 1.7 points of perplexity on both the validation and test sets, while we did not make any change to the architecture of the AWD-LSTM language model.

\subsection{Wikitext-103}

The current state-of-the-art language model trained on Wikitext-103 dataset is the Transformer-XL \cite{dai2018transformer}. We apply our method on the state-of-the-art Transformer-XL Large model, which has 18 layers and 257M parameters. The input size and hidden size are 1024. 16 attention heads are used. We regard the first 14 layers as the phrase generator and the last 4 layers as the word generator. In other words, the context-phrase alignment is trained with the outputs of the 14th layer.


\begin{figure*}[t!]
    \centering
    \begin{subfigure}[t]{0.5\textwidth}
        \centering
        \includegraphics[height=2in]{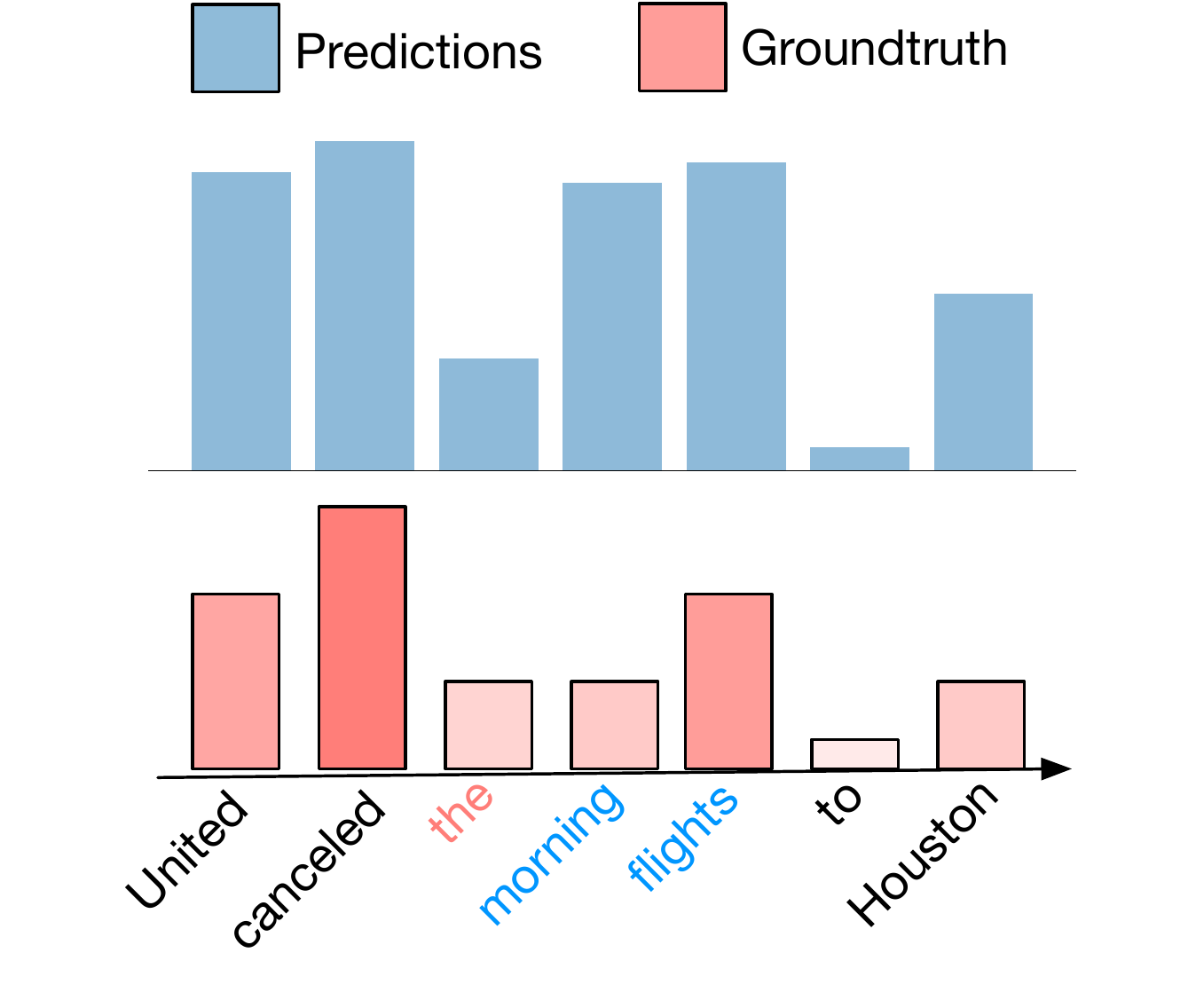}
        \caption{Syntactic heights of each word.}
        \label{fig:e1:a}
    \end{subfigure}%
    ~ 
    \begin{subfigure}[t]{0.5\textwidth}
        \centering
        \includegraphics[height=2in]{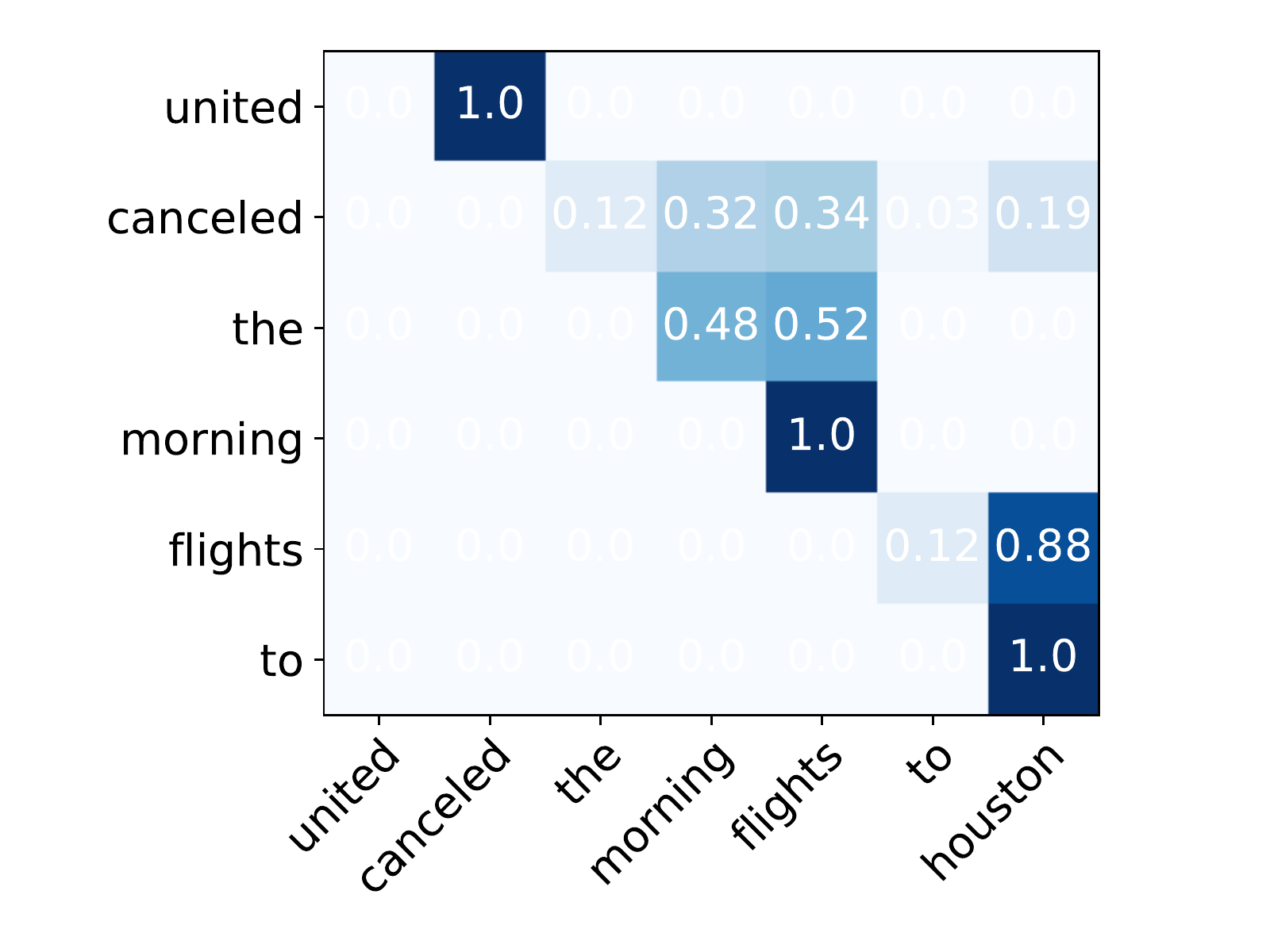}
        \caption{Induced phrases and headword attentions.}
        \label{fig:e1:b}
    \end{subfigure}
    \caption{Examples of induced phrases and corresponding headword attention for generating the phrase embedding. The word of each row stands for the target word as the current input of the language model, and the values in each row in the matrices stands for the words consisting the induced phrase and their weights.}
    \label{fig:e1}
\end{figure*}

The model is trained on 4 Titan X Pascal GPUs, each of which has 12G memory. Because of the limitation of computational resources, we use our approach to fine-tune the officially released pre-trained Transformer-XL Large model for 1 epoch. The experimental results are shown in Table \ref{tab:wt103}. Our approach got 17.4 perplexity with the officially released evaluation scripts, significantly outperforming all baselines and achieving new state-of-the-art performance\footnote{We did not show Dev PPLs in Table \ref{tab:wt103} since only the correct approach to reproduce the test PPL was provided with the pretrained Transformer-XL model.}.

\section{Discussion}

\begin{figure*}[t!]
    \centering
    \begin{subfigure}[t]{0.45\textwidth}
        \centering
        \includegraphics[height=1.8in]{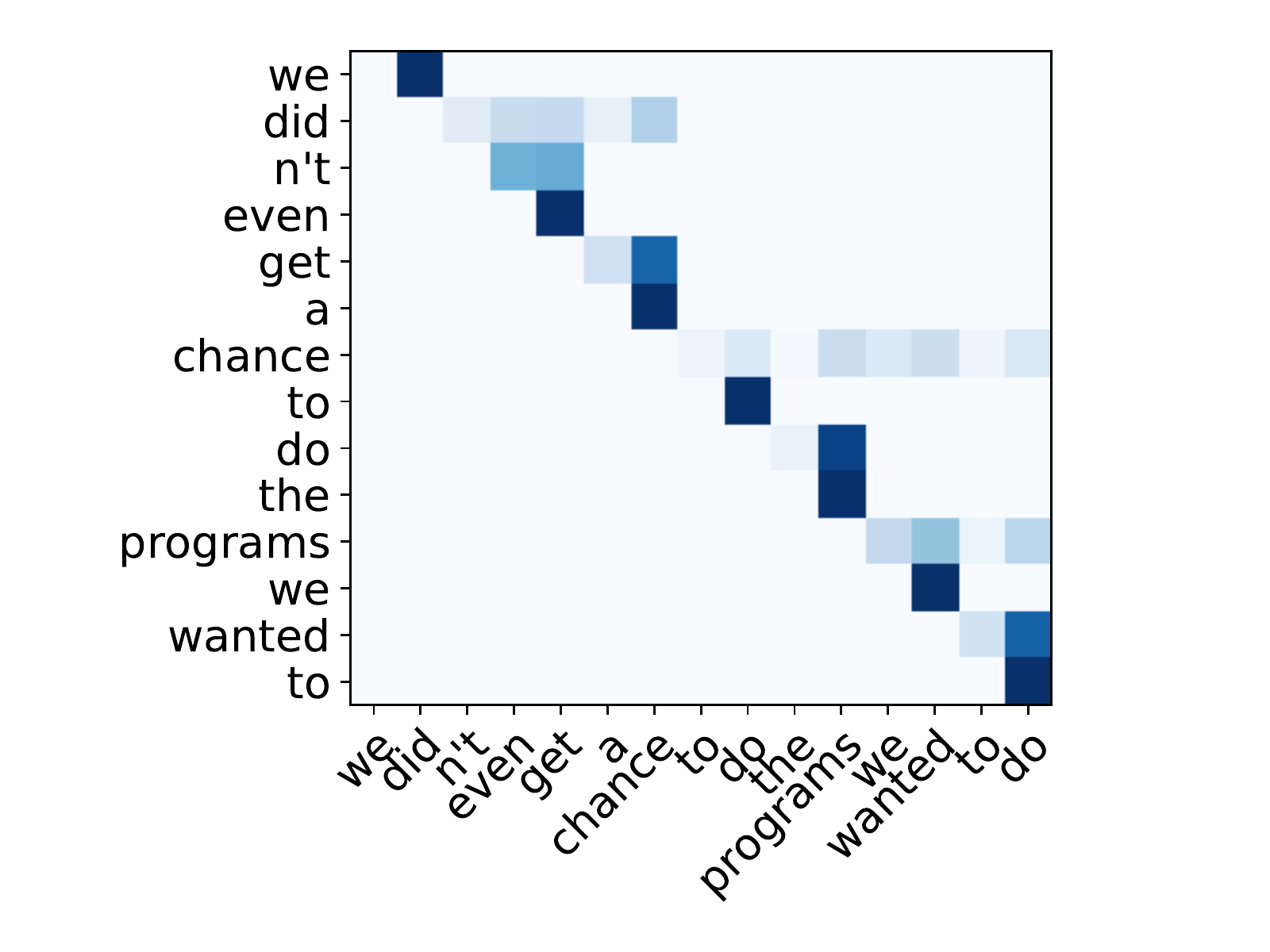}
        \caption{}
        \label{fig:e2:a}
    \end{subfigure}
    ~ 
    \begin{subfigure}[t]{0.45\textwidth}
        \centering
        \includegraphics[height=1.8in]{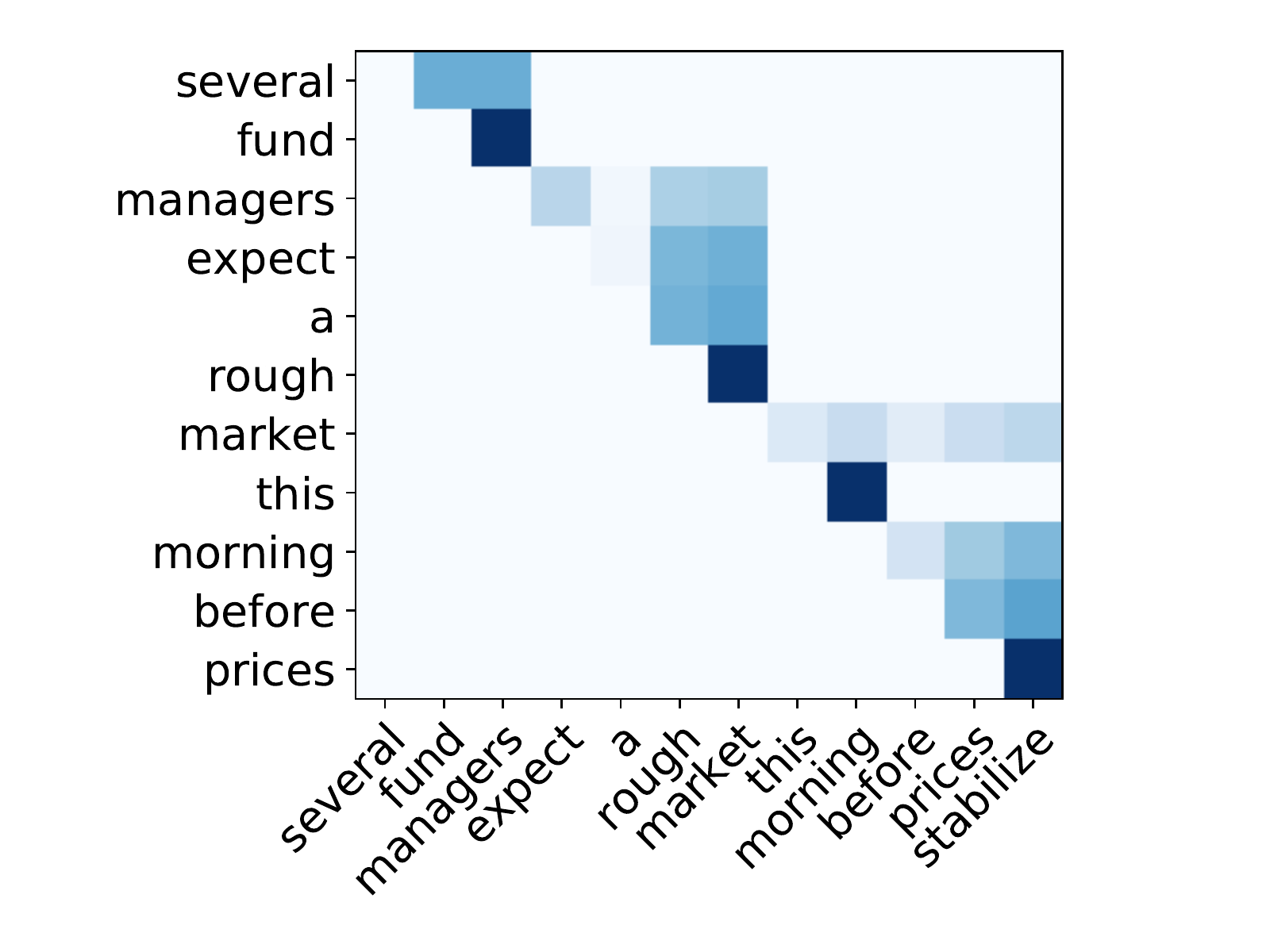}
        \caption{}
        \label{fig:e2:b}
    \end{subfigure}
    ~
    \begin{subfigure}[t]{0.45\textwidth}
        \centering
        \includegraphics[height=1.8in]{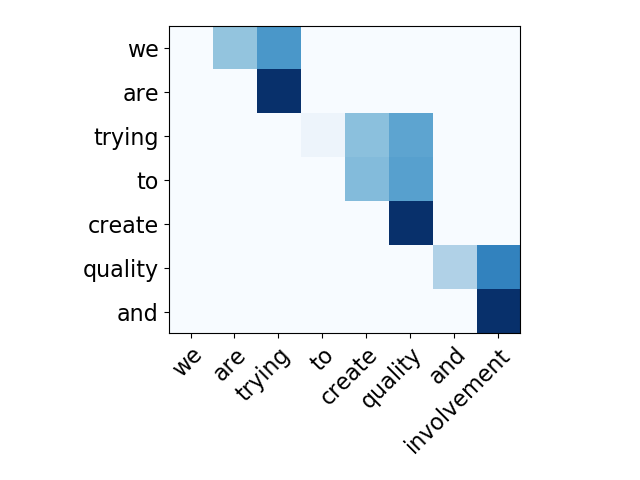}
        \caption{}
        \label{fig:e2:c}
    \end{subfigure}
    ~
    \begin{subfigure}[t]{0.45\textwidth}
        \centering
        \includegraphics[height=1.8in]{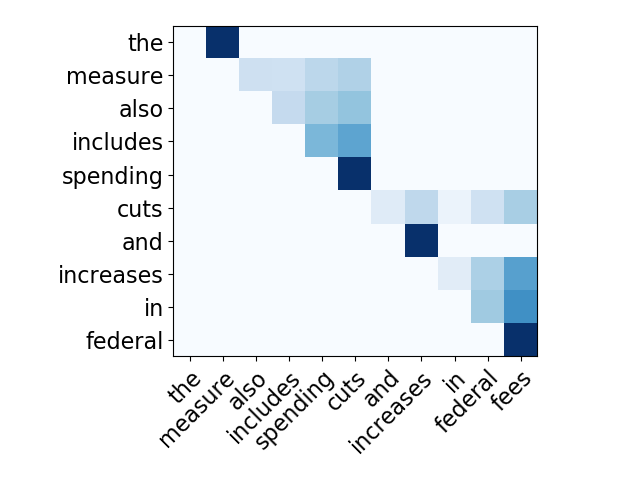}
        \caption{}
        \label{fig:e2:d}
    \end{subfigure}
    ~
    \begin{subfigure}[t]{0.45\textwidth}
        \centering
        \includegraphics[height=1.8in]{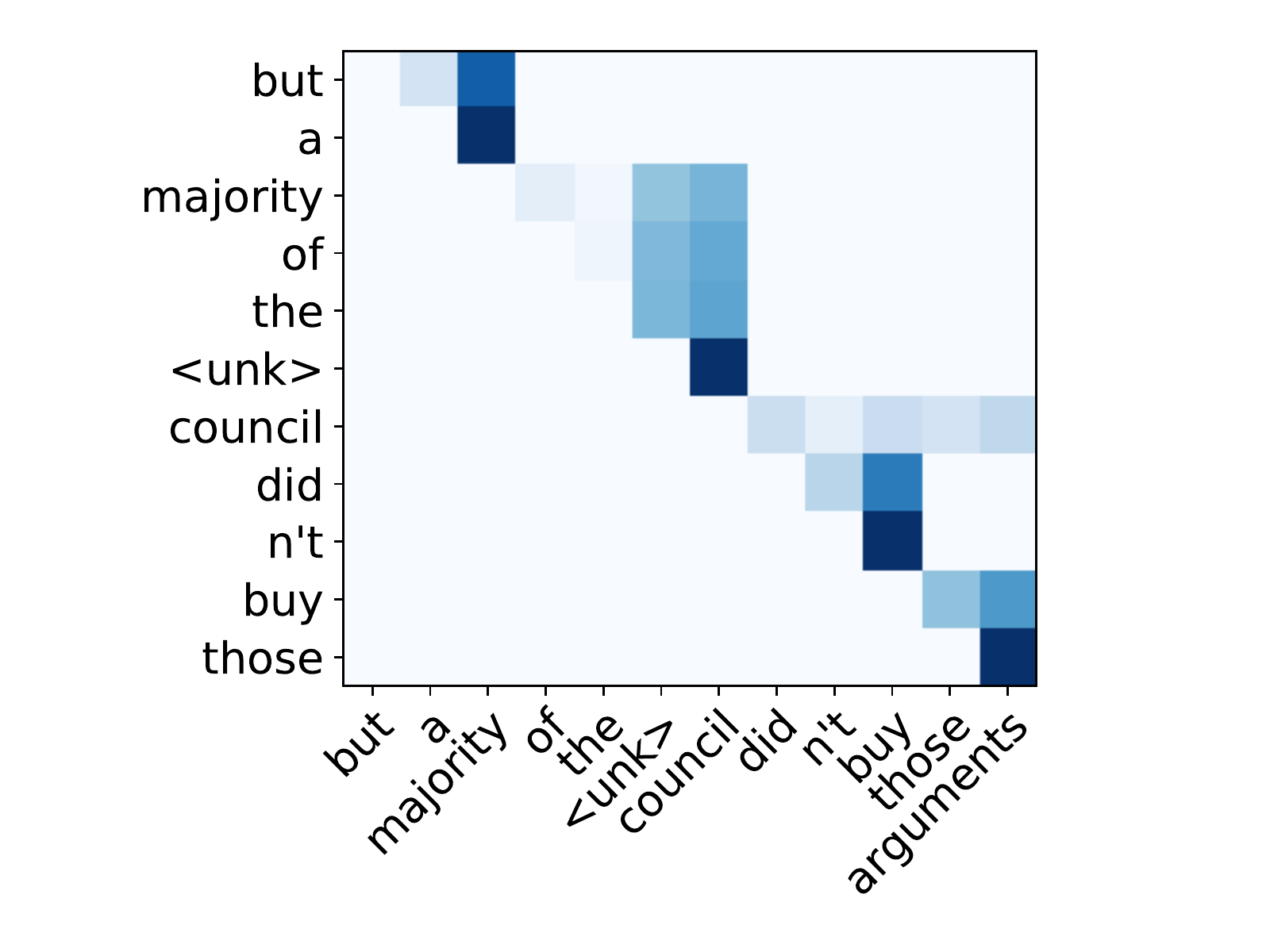}
        \caption{}
        \label{fig:e2:e}
    \end{subfigure}
    ~
    \begin{subfigure}[t]{0.45\textwidth}
        \centering
        \includegraphics[height=1.8in]{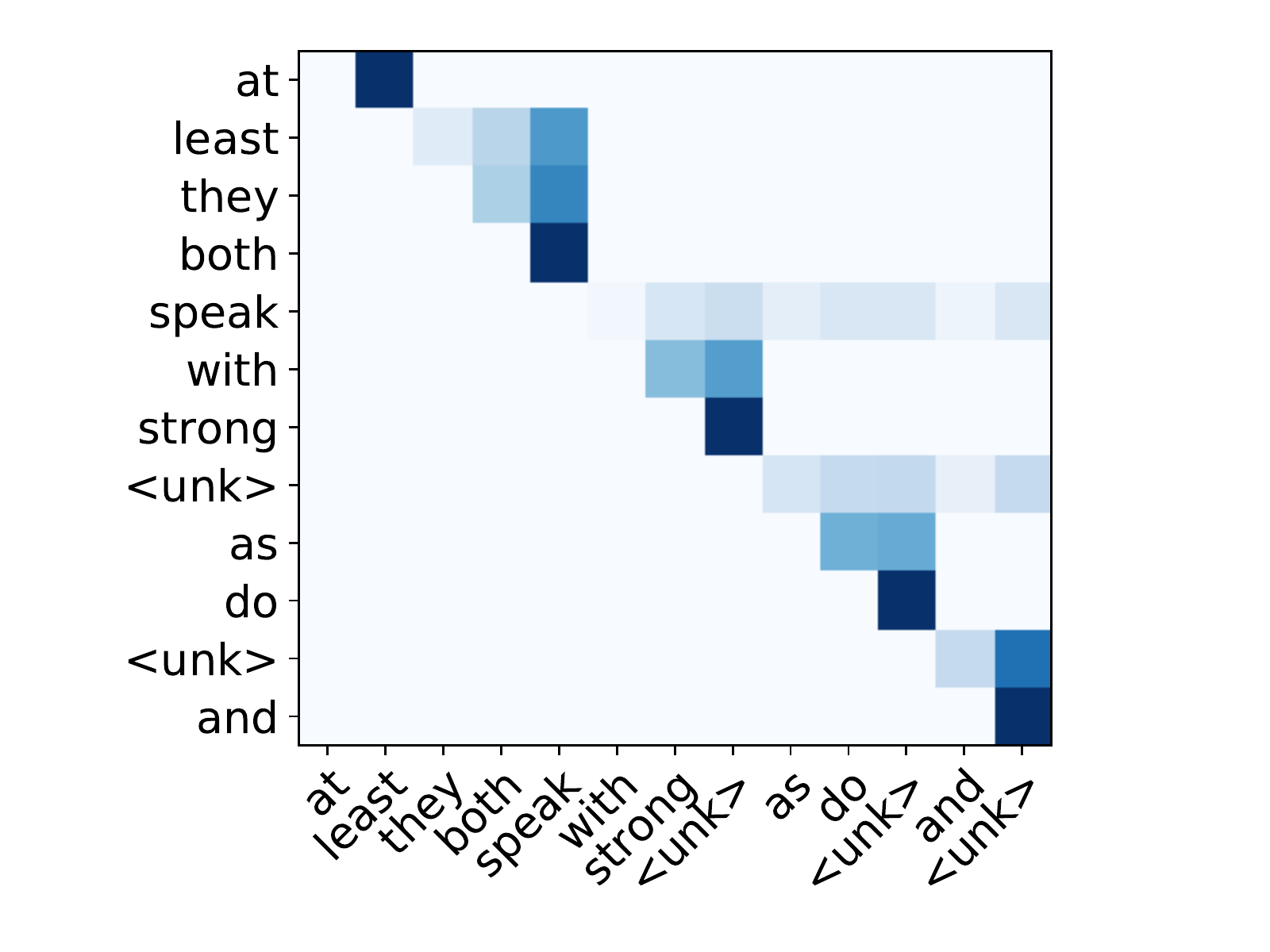}
        \caption{}
        \label{fig:e2:f}
    \end{subfigure}
    \caption{Examples of phrase inducing and headword attentions.}
    \label{fig:e2}
\end{figure*}

In this section, we show what is learned by training language models with the context-phrase alignment objective function by visualizing the syntactic heights output by the TCN model and the phrases induced by each target word in a sentence. We also visualize the headword attentions over the induced phrase.

The first example is the sentence showed in Figure \ref{fig:dt}. The sentence came from \citet{jurafsky2014speech} and did not appear in our training set.  Figure \ref{fig:dt} shows the syntactic heights and the induced phrase of ``the'' according to the ground-truth dependency information. Our model is not given such high-quality inputs in either training or evaluation.


Figure \ref{fig:e1} visualizes the structure learned by our phrase induction model. The inferred syntactic heights are shown in Figure \ref{fig:e1:a}. Heights assigned to words ``the'' and ``to'' are significantly lower than others, while the verb ``canceled'' is assigned the highest in the sentence. Induced phrases are shown in Figure \ref{fig:e1:b}. The words at the beginning of each row stand for the target word of each step. Values in the matrix stand for attention weights for calculating phrase embedding. The weights are calculated with the phrase segmenting conditions (PSC) and the syntactic heights described in Equations \ref{eq:psc-1} to \ref{eq:att}. For the target word ``united'', $h_{united} < h_{canceled}$ and $h_{canceled} > h_{the}$, hence the induced phrase of ``united'' is a single word ``canceled'', and the headword attention of ``canceled'' is 1, which is indicated in the first row of Figure \ref{fig:e1:b}. The phrase induced by ``canceled'' is the entire following sequence, ``the morning flights to houston'', since no following word has a higher syntactic height than the target word. It is also shown that the headword of the induced phrase of ``canceled'' is ``flights'', which agrees with the dependency structure indicated in Figure \ref{fig:dt}.

More examples are shown in Figure \ref{fig:e2}. Figures \ref{fig:e2:a} to \ref{fig:e2:d} show random examples without any unknown word, while the examples shown in Figures \ref{fig:e2:e} and \ref{fig:e2:f} are randomly selected from sentences with unknown words, which are marked with the UNK symbol. The examples show that the phrase induction model does not always predict the exact structure represented by the dependency tree. For example, in Figure \ref{fig:e2:b}, the TCN model assigned the highest syntactic height to the word ``market'' and induced the phrase ``expect a rough market'' for the context ``the fund managers''. However, in a ground-truth dependency tree, the verb ``expect'' is the word directly connected to the root node and therefore has the highest syntactic height.

Although not exactly matching linguistic dependency structures, the phrase-level structure predictions are reasonable. The segmentation is interpretable and the predicted headwords are appropriate. In Figure \ref{fig:e2:c}, the headwords are ``trying'', ``quality'', and ``involvement''. The model is also robust with unknown words. In Figure \ref{fig:e2:e}, ``the $<$unk$>$ council'' is segmented as the induced phrase of ``but a majority of''. In this case, the model recognized that the unknown word is dependent on ``council''.

The sentence in Figure \ref{fig:e2:f} includes even more unknown words. However, the model still correctly predicted the root word, the verb ``speak''. For the target word ``with'', the induced phrase is ``strong $<$unk$>$''. Two unknown words are located in the last few words of the sentence. The model failed to induce the phrase ``$<$unk$>$ and $<$unk$>$'' for the word ``do'', but still successfully split ``$<$unk$>$'' and ``and''. Meanwhile, the attentions over the phrases induced by ``speak'', ``do'', and the first ``$<$unk$>$'' are not quite informative, suggesting that unknown words made some difficulties for headword prediction in this example. However, the unknown words are assigned significantly higher syntactic heights than the word ``and''.

\section{Conclusion}

In this work, we improved state-of-the-art language models by aligning context and induced phrases. We defined syntactic heights and phrase segmentation rules. The model generates phrase embeddings with headword attentions. We improved the AWD-LSTM and Transformer-XL language models on different data sets and achieved state-of-the-art performance on the Wikitext-103 corpus. Experiments showed that our model successfully learned approximate phrase-level knowledge, including segmentation and headwords, without any annotation. In future work, we aim to capture better structural information and possible connections to unsupervised grammar induction. 


\bibliography{acl2019}
\bibliographystyle{acl_natbib}

\end{document}